\documentclass{article}

\usepackage{PRIMEarxiv}

\usepackage[utf8]{inputenc} 
\usepackage[T1]{fontenc}    
\usepackage{hyperref}       
\usepackage{url}            
\usepackage{booktabs}       
\usepackage{amsfonts}       
\usepackage{nicefrac}       
\usepackage{microtype}      
\usepackage{lipsum}
\usepackage{fancyhdr}       
\usepackage{graphicx}       
\usepackage{amsmath}
\usepackage{bm}
\usepackage[ruled,vlined,linesnumbered]{algorithm2e}
\graphicspath{{media/}}     

\title{"Generative Models for Financial Time Series Data: Enhancing Signal-to-Noise Ratio and Addressing Data Scarcity in A-Share Market
}

\author{
  Guangming CHE \\
    noeche@sjtu.edu.cn \\
}

\begin{document}
\maketitle

\begin{abstract}
The financial industry is increasingly seeking robust methods to address the challenges posed by data scarcity and low signal-to-noise ratios, which limit the application of deep learning techniques in stock market analysis. This paper presents two innovative generative model-based approaches to synthesize stock data, specifically tailored for different scenarios within the A-share market in China. The first method, a sector-based synthesis approach, enhances the signal-to-noise ratio of stock data by classifying the characteristics of stocks from various sectors in China's A-share market. This method employs an Approximate Non-Local Total Variation algorithm to smooth the generated data, a bandpass filtering method based on Fourier Transform to eliminate noise, and Denoising Diffusion Implicit Models to accelerate sampling speed. The second method, a recursive stock data synthesis approach based on pattern recognition, is designed to synthesize data for stocks with short listing periods and limited comparable companies. It leverages pattern recognition techniques and Markov models to learn and generate variable-length stock sequences, while introducing a sub-time-level data augmentation method to alleviate data scarcity issues.

We validate the effectiveness of these methods through extensive experiments on various datasets, including those from the main board, STAR Market, Growth Enterprise Market Board, Beijing Stock Exchange, NASDAQ, NYSE, and AMEX. The results demonstrate that our synthesized data not only improve the performance of predictive models but also enhance the signal-to-noise ratio of individual stock signals in price trading strategies. Furthermore, the introduction of sub-time-level data significantly improves the quality of synthesized data, particularly in scenarios with limited comparable companies and short listing periods. This research contributes to the financial data synthesis domain by providing new tools and techniques that can support financial analysis and high-frequency trading, offering valuable insights into the complex dynamics of the A-share market.
\end{abstract}

\keywords{Financial Data Synthesis \and Signal-to-Noise Ratio Enhancement \and Data Scarcity Mitigation \and Generative Models for A-Share Market}

\section{Introduction}
Introduction

The realm of finance is rife with complexities and uncertainties that make the accurate prediction of financial instrument prices and returns a formidable task. Such predictions are essential for guiding investment decisions, optimizing asset allocation, and managing risk effectively. The accuracy of these predictions, however, is often hindered by the limitations in data quality and quantity, particularly in the stock market where issues like low signal-to-noise ratios and data homogeneity pose significant challenges to constructing high-precision predictive models \cite{Li2022}.

Financial data is not only sensitive but also of high value, and its mishandling can lead to severe security risks for both companies and users \cite{Li2022}. The implementation of data privacy regulations has further exacerbated the challenges faced by the financial industry in acquiring and sharing data, resulting in information asymmetry and isolated data silos \cite{Li2022}. In response to these challenges, the advent of artificial intelligence and deep learning models offers innovative solutions for generating synthetic financial data. These methods aim to retain the characteristics of original data, increase data diversity, protect privacy, and enhance the training and predictive accuracy of financial models \cite{Gao2024}.

Despite the vastness and complexity of financial market data, underlying structures and patterns can be revealed through in-depth research and analysis \cite{Rama2001}. Financial data distribution is characterized by traits such as leptokurtosis, heteroskedasticity, and volatility clustering, which are crucial for understanding market dynamics \cite{Rama2001}. These characteristics make the task of generating synthetic financial data not only essential but also challenging, as it requires capturing the subtleties and variabilities inherent in financial time series \cite{Huang2024}.

This study addresses the limitations of existing stock data generation methods, particularly their inability to fully consider the unique rules and characteristics of China's A-share market. We propose two innovative methods for synthesizing stock data that can enhance the signal-to-noise ratio and address data scarcity issues, especially for stocks with short listing times and limited comparable companies. Our approach leverages generative models to create synthetic financial data that can improve the performance of predictive models and provide valuable insights into market trends and behaviors \cite{Huang2024}.

\section{Background and Related Work}

\subsection{Financial Market Dynamics and Data Scarcity}
The financial markets are intricate systems where the accurate prediction of asset prices and returns is crucial for investment strategies and risk management. However, the inherent volatility and unpredictability of these markets make accurate forecasting a significant challenge. A key issue in this context is the scarcity and low quality of financial data, which limits the application of deep learning techniques within the financial industry. Particularly in stock markets, the low signal-to-noise ratio and high data homogeneity hinder the construction of precise predictive models \cite{Li2022}.

The sensitivity and high value of financial data also mean that any leakage or malicious manipulation can pose serious security risks to both financial institutions and their customers \cite{Li2022}. With the enforcement of data privacy regulations, the financial sector is facing increased difficulties in data acquisition and sharing, leading to information asymmetry and the creation of "data silos" \cite{Li2022}. These challenges have spurred the exploration of artificial intelligence techniques to generate synthetic financial data that can maintain the characteristics of original data, increase data diversity, protect privacy, and enhance the effectiveness of model training and prediction accuracy \cite{Gao2024}.

\subsection{Generative Models in Finance}
Generative models, such as Variational Autoencoders (VAEs), Generative Adversarial Networks (GANs), and diffusion-based models, have shown great potential in generating synthetic data across various domains, including finance. These models have been used to create synthetic financial data that mimics the statistical properties of real-world market behavior and customer trading patterns \cite{Kingma2013, GAN2014, ddpm2020}.

VAEs, introduced by Kingma et al. in 2013 \cite{Kingma2013}, have been particularly successful in generating new data instances similar to the training data by learning to encode and decode data in a probabilistic manner. GANs, proposed by Goodfellow et al. in 2014 \cite{GAN2014}, have revolutionized the field of image generation by employing a competitive approach between a generator and a discriminator. Diffusion models, initially proposed by Sohl-Dickstein et al. in 2015 \cite{DM2015} and later improved by Ho and Jain in 2020 \cite{ddpm2020}, have demonstrated their ability to generate high-quality samples by gradually reversing the noise injection process.

\subsection{Challenges in Financial Data Generation}
Despite the advancements in generative models, their application in financial data generation is still in its nascent stages. The complexity and dynamism of financial markets present numerous challenges. Financial data generation requires a deep understanding of market dynamics and the exploration of how to effectively integrate deep generative models with the characteristics of financial data to produce high-quality, practical synthetic data \cite{Huang2024}.

Moreover, these models are often developed and trained on data from Western markets, such as the U.S. stock market, and may not fully account for the unique rules and characteristics of markets like China's A-share market. For instance, the distribution and regulatory differences between these markets can lead to synthetic data that does not conform to the rules and patterns of specific financial markets \cite{jpm2019, jpm2021}.

\subsection{Financial Data Synthesis Approaches}
To address these challenges, various financial institutions and scholars have proposed frameworks for generating synthetic financial data. These frameworks aim to create datasets that are statistically similar to real data but without revealing any entity information, thus supporting financial analysis and research while maintaining privacy constraints \cite{jpm2019}. Academic research has also explored different variations of GANs to improve performance in financial data synthesis, including architectural variants and loss function variants \cite{Takahashi2019, cGAN-fin2019, pardo2019}.

In conclusion, the background and related work highlight the importance of addressing data scarcity and the unique challenges of financial data generation. The exploration of generative models in finance is still an active area of research, with significant potential for developing more sophisticated methods to generate synthetic data that can enhance the predictive power of financial models and strategies.

\section{Score-based Generative Models}

Score-based Generative Models (SGMs) are a type of self-supervised machine learning method used for generating new data samples. The core idea of SGMs is to learn the score function of an unknown data distribution $p_{\text{data}}(\bm{x})$, which is the gradient of the log-probability density function with respect to the data $\bm{x}$, $\nabla_{\bm{x}} \log p(\bm{x})$, to guide the generation of new samples. Once we have an estimate of this score function, we can start from a random data point and move it to a position with higher probability density using gradient ascent.

In practice, a neural network $s_\theta$ is used to approximate the score $\nabla_{\bm{x}} \log p(\bm{x})$, and the goal of training this neural network is to minimize the following objective function:
\begin{equation}
    L(\theta) = \frac{1}{2} \mathbb{E}_{p_{\text{data}}(\bm{x})} \left[ \|s_\theta(\bm{x}) - \nabla_{\bm{x}} \log p_{\text{data}}(\bm{x})\|^2_2 \right] = \mathbb{E}_{p_{\text{data}}(\bm{x})} \left[ \operatorname{tr}\left(\nabla_{\bm{x}} s_\theta(\bm{x})\right) + \frac{1}{2} \|s_\theta(\bm{x})\|^2_2 \right].
\end{equation}
According to Score Matching, $L(\theta)$ can be written in the following form\cite{scorematch}, and the derivation is detailed in Appendix A:
\begin{equation}
    L(\theta) = \mathbb{E}_{p_{\text{data}}(\bm{x})} \left[ \operatorname{tr}\left(\nabla_{\bm{x}} s_\theta(\bm{x})\right) + \frac{1}{2} \|s_\theta(\bm{x})\|^2_2 \right].
\end{equation}
However, directly computing the $\operatorname{tr}(\nabla_{\bm{x}} s_\theta(\bm{x}))$ in the objective function requires a large amount of computational resources. Therefore, Vincent et al. proposed Denoising Score Matching (DSM) as a solution to avoid this computation\cite{Vincent2011}. DSM perturbs the data $\bm{x}$ with a given noise distribution $q_\sigma (\tilde{\bm{x}}|\bm{x})$ and then attempts to estimate the score of the perturbed data distribution $q_\sigma (\tilde{\bm{x}})$. At this point, the training goal becomes minimizing the following objective function:
\begin{equation}
    \frac{1}{2} \mathbb{E}_{q_\sigma (\tilde{\bm{x}}|\bm{x})p_{\text{data}}(\bm{x})} \left[ \|s_\theta(\tilde{\bm{x}}) - \nabla_{\tilde{\bm{x}}} \log q_\sigma (\tilde{\bm{x}}|\bm{x})\|^2_2 \right].
\end{equation}

In some cases, the training goal of DSM is equivalent to the training goal of Denoising Autoencoders (DAE), which also follows a simple noise-adding and denoising process.

These models and methods can be unified through Stochastic Differential Equations (SDEs), which provide a framework for understanding how to gradually add noise from a known initial distribution and how to remove noise from a perturbed distribution to recover the original data distribution. This offers a powerful tool for designing and analyzing score-based generative models and enables different variants and extensions.

\subsection{Score Matching with Langevin Dynamics}

The process of Score Matching with Langevin Dynamics (SMLD) includes two parts. On the one hand, it uses denoising score matching to estimate the score of the perturbed data distribution; on the other hand, it uses Langevin dynamics to iteratively sample from the prior distribution.

Song et al. proposed perturbing the data with multiple levels of noise and training a Noise Conditioned Score Network (NCSN)\cite{songyang2019} to estimate the score corresponding to all noise levels. The perturbation method is defined as:
\[ q_\sigma (\tilde{\bm{x}} | \bm{x}) = \mathcal{N}(\tilde{\bm{x}} | \bm{x}, \sigma^2 I), \]
where $I$ is the identity matrix. The probability distribution of the perturbed data is:
\[ q_\sigma (\tilde{\bm{x}}) = \int p_{\text{data}}(\bm{x}) \mathcal{N}(\tilde{\bm{x}} | \bm{x}, \sigma^2 I) d\bm{x}. \]

Song et al. also defined a noise sequence $\{\sigma_i\}_{i=1}^{N}$ that satisfies the following conditions: $\sigma_{\min} = \sigma_1 < \sigma_2 < \cdots < \sigma_N = \sigma_{\max}$, where $\sigma_{\min}$ is small enough to make $q_{\sigma_{\min}}(\tilde{\bm{x}}) \approx p_{\text{data}}(\bm{x})$, and $\sigma_{\max}$ is large enough to make $q_{\sigma_{\max}}(\tilde{\bm{x}}) \approx \mathcal{N}(\bm{x} | 0, \sigma_{\max}^2 I)$. To estimate the score of the perturbed data distribution, a conditional score network $s_\theta(\tilde{\bm{x}}, \sigma)$ is trained to satisfy for all $\sigma \in \{\sigma_i\}_{i=1}^{N}$:
\[ s_\theta(\tilde{\bm{x}}, \sigma) \approx \nabla_{\bm{x}} \log q_\sigma(\tilde{\bm{x}}). \]
The training goal of NCSN is the weighted sum of the denoising score matching objectives, that is, to find the optimal parameters $\theta^*$ to minimize the following loss function:
\[ \theta^* = \arg\min_\theta \sum_{i=1}^{N} \sigma_i^2 \mathbb{E}_{p_{\text{data}}(\bm{x})} \mathbb{E}_{q_{\sigma_i} (\tilde{\bm{x}} | \bm{x})} \left[ \|s_\theta(\tilde{\bm{x}}, \sigma_i) - \nabla_{\tilde{\bm{x}}} \log q_{\sigma_i} (\tilde{\bm{x}} | \bm{x})\|^2_2 \right]. \]

SMLD employs a sampling method based on Langevin MCMC to generate new data samples. The update rule for this method is as follows:
\[ x_i^{(m)} = x_i^{(m-1)} + \epsilon_i s_\theta^*(x_i^{(m-1)}, \sigma_i) + \sqrt{2\epsilon_i} z_i^{(m)}, \quad m = 1, 2, \ldots, M, \]
where: $\epsilon_i$ is the step size, controlled by the noise level sequence $\{\sigma_i\}_{i=1}^{N}$; $z_i^{(m)}$ is a standard Gaussian variable. SMLD needs to iterate according to $i = N, N-1, \ldots, 1$, starting from $x_0^{(N)} \sim \mathcal{N}(\bm{x} \mid 0, \sigma_{\max}^2 I)$, and setting $x_0^{(i)} = x_M^{(i+1)}$. When $M \to \infty$ and $\epsilon_i \to 0$, $x_M^{(1)}$ will become an exact sample from $q_{\sigma_{\min}}(\tilde{\bm{x}}) \approx p_{\text{data}}(\bm{x})$.

\subsection{Denoising Diffusion Probabilistic Model}
The Denoising Diffusion Probabilistic Model (DDPM)\cite{ddpm2020} can be regarded as a type of hierarchical Markov variational autoencoder\cite{luo2022}. Consider a noise sequence \(0 < \beta_1 < \beta_2 < \cdots < \beta_N < 1\) and the forward noise process \(p(\bm{x}_i | \bm{x}_{i-1}) = \mathcal{N}(\bm{x}_i | \sqrt{1 - \beta_i} \bm{x}_{i-1}, \beta_i I)\). Define \(\alpha_i = \prod_{j=1}^{i} (1 - \beta_j)\), then we have:
\begin{equation}
    p_{\alpha_i} (\bm{x}_i | \bm{x}_0) = \mathcal{N}(\bm{x}_i | \sqrt{\alpha_i} \bm{x}_0, (1 - \alpha_i) I).
\end{equation}
Similar to Score Matching with Langevin Dynamics (SMLD), the perturbed data distribution can be expressed as:
\begin{equation}
    p_{\alpha_i} (\tilde{\bm{x}}) = \int p_{\text{data}}(\bm{x}) p_{\alpha_i} (\tilde{\bm{x}} | \bm{x}) d\bm{x}.
\end{equation}
Here, the noise scale is preset to satisfy \(p_{\alpha_N} (\tilde{\bm{x}}) \sim \mathcal{N}(0, I)\), meaning that the final perturbed data is close to a standard Gaussian distribution. According to the reparameterization method\cite{ddpm2020}, we can obtain the conditional distribution \(q(\bm{x}_t | \bm{x}_0)\):
\begin{equation}
    q(\bm{x}_t | \bm{x}_0) = \mathcal{N} \left( \bm{x}_t; \sqrt{\overline{\alpha_t}} \bm{x}_0, (1 - \overline{\alpha_t}) I \right),
\end{equation}
where \(\overline{\alpha_t} = \prod_{i=1}^{t} \alpha_i\) and \(\alpha_t = 1 - \beta_t\). Then, \(\bm{x}_t\) can be written as:
\begin{equation}
    \bm{x}_t = \sqrt{\overline{\alpha_t}} \bm{x}_0 + \sqrt{1 - \overline{\alpha_t}} \epsilon,
\end{equation}
where \(\epsilon \sim \mathcal{N}(0, I)\). The reverse denoising process can be written as:
\begin{equation}
    p_\theta (\bm{x}_{i-1} | \bm{x}_i) = \mathcal{N}\left(\bm{x}_{i-1}; \frac{1}{\sqrt{1 - \beta_i}} \left( \bm{x}_i + \beta_i s_\theta (\bm{x}_i, i) \right), \beta_i I\right).
\end{equation}

The training objective of DDPM is the sum of the weighted Evidence Lower Bound (ELBO), that is, to find the optimal parameters \(\theta^*\) to minimize the following loss function:
\begin{equation}
    \theta^* = \arg\min_\theta \sum_{i=1}^{N} (1 - \alpha_i) \mathbb{E}_{p_{\text{data}}(\bm{x})} \mathbb{E}_{p_{\alpha_i} (\tilde{\bm{x}} | \bm{x})} \left[ \|s_\theta^*(\tilde{\bm{x}}, i) - \nabla_{\tilde{\bm{x}}} \log p_{\alpha_i} (\tilde{\bm{x}} | \bm{x})\|^2_2 \right].
\end{equation}
where \(\alpha_i = \prod_{j=1}^{i} (1 - \beta_j)\) is the cumulative product, representing the total noise influence from the initial data to the \(i\)-th step; \(p_{\alpha_i} (\tilde{\bm{x}} | \bm{x}) = \mathcal{N}(\tilde{\bm{x}} | \sqrt{\alpha_i} \bm{x}, (1 - \alpha_i) I)\) is the conditional distribution of the perturbed data \(\tilde{\bm{x}}\) given the original data \(\bm{x}\).

\subsection{Unifying SMLD and DDPM from the Perspective of Stochastic Differential Equations}

Based on the research by Song et al. \cite{Song2020ScoreBasedGM}, Score Matching with Langevin Dynamics (SMLD) and Denoising Diffusion Probabilistic Models (DDPM) can be unified under the perspective of Stochastic Differential Equations (SDEs). Let ${\bm{x}(t)}_{t=0}^{T}$ be a stochastic diffusion process indexed by a continuous time variable $t \in [0, T]$. Let $p_0$ represent the true data distribution, and $p_T$ be a tractable prior distribution such that $\bm{x}_0 \sim p_0$ and $\bm{x}_T \sim p_T$.

Using $p_t(\bm{x})$ to denote the probability density function of $\bm{x}(t)$, and $p_{st}(\bm{x}(t) \mid \bm{x}(s))$ to represent the transition kernel from $\bm{x}(s)$ to $\bm{x}(t)$, where $0 \leq s < t \leq T$. A stochastic differential equation can be used to represent this forward diffusion process:
\begin{equation}
d\bm{x} = f(\bm{x}, t) dt + g(t) d\bm{w},
\end{equation}
where $f(\bm{x}, t) dt$ is the drift term, representing the average trend of the system's change; $g(t) d\bm{w}$ is the diffusion term, representing the strength of random fluctuations; here, $\bm{w}$ is a standard Wiener process, and $d\bm{w} \sim \mathcal{N}(0, I)$, meaning that $d\bm{w}$ follows a Gaussian distribution with mean zero and covariance matrix equal to the identity matrix $I$, representing unpredictable random noise. The synthetic data generation process is the reverse process of (3.11), which is also a stochastic differential equation:
\begin{equation}
d\bm{x} = \left[ f(\bm{x}, t) - g^2(t) \nabla_{\bm{x}} \log p_t (\bm{x}) \right] dt + g(t) d\bar{\bm{w}},
\end{equation}
where $\bar{\bm{w}}$ is the reverse-time Wiener process; $\nabla_{\bm{x}} \log p_t (\bm{x})$ is the score corresponding to the marginal distribution $p_t (\bm{x})$ at each $t$. This process starts from the initial data point $\bm{x}T \sim p_T$, gradually denoises, and finally generates a sample $\bm{x}0$ close to the true data distribution $p_0$. In theory, if $dt$ is small enough, we can gradually obtain $\bm{x}0 \sim p_0$ through (3.12). To estimate $\nabla{\bm{x}} \log p_t (\bm{x})$, the score of the marginal distribution, a score network $s\theta(\bm{x}, t)$ is trained, with the objective function being:
\begin{equation}
\lambda(t) \mathbb{E}t \mathbb{E}{\bm{x}0} \mathbb{E}{\bm{x}t \mid \bm{x}0} \left[ |s\theta(\bm{x}(t), t) - \nabla{\bm{x}(t)} \log p{0t} (\bm{x}(t) \mid \bm{x}(0))|^2_2 \right],
\end{equation}
where $\lambda : [0, T] \to \mathbb{R}^+$ is a positive weight function; $t \sim U[0, T]$, indicating that $t$ is drawn from a uniform distribution $U[0, T]$.

DDPM can be seen as a discrete form of continuous-time SDEs. Typically, the discrete forward process for SMLD is:
\begin{equation}
\bm{x}i = \bm{x}{i-1} + \sqrt{\sigma_i^2 - \sigma_{i-1}^2} \bm{z}{i-1}, \quad i = 1, \cdots, N.
\end{equation}
As $N \to \infty$, the discrete form ${\bm{x}i}{i=1}^{N}$ becomes the continuous form ${\bm{x}(t)}{t=0}^{1}$, and the continuous forward process can be written as:
\begin{equation}
d\bm{x} = \sqrt{\frac{d[\sigma^2(t)]}{dt}} d\bm{w},
\end{equation}
where $f(\bm{x}, t) = 0$ and $g(t) = \sqrt{\frac{d[\sigma^2(t)]}{dt}}$. This is called the Variance Exploding SDE (VE-SDE).

For DDPM, the discrete forward process is:
\begin{equation}
\bm{x}i = \sqrt{1 - \beta_i} \bm{x}{i-1} + \sqrt{\beta_i} \bm{z}_{i-1}, \quad i = 1, \cdots, N.
\end{equation}
As $N \to \infty$, the continuous form of the DDPM forward process can be written as:
\begin{equation}
d\bm{x} = -\frac{\beta(t)}{2} \bm{x} dt + \sqrt{\beta(t)} d\bm{w},
\end{equation}
where $f(\bm{x}, t) = -\frac{\beta(t)}{2} \bm{x}$ and $g(t) = \sqrt{\beta(t)}$. This is called the Variance Preserving SDE (VP-SDE).

By replacing $f(\bm{x}, t)$ and $g(t)$ in (3.12) with those in (3.15) or (3.17), we can obtain the SDE form of the reverse process corresponding to SMLD or DDPM.

\section{Methodology Design}

The task of stock prediction is challenging, and the scarcity of data is a significant reason. To fully leverage the potential of machine learning models, ample and high-quality data is crucial. However, obtaining high-quality stock data within a specific target domain is often very difficult. In this study, we utilize diffusion models (DMs) to propose a new method for synthesizing A-share data—CS-Diffusion. This method generates additional data points to augment stock data, thereby overcoming the problem of data scarcity and enabling us to more accurately predict the potential return rates (RR) of stocks in the real world.

\subsection{Training Method Based on A-Share Market Plate Type}
\subsubsection{Problem Definition}

Let $\bm{x}$ and $\bm{c}$ be two time series of length $L$, where $\bm{x} \in \mathbb{R}^L$ is the input to the diffusion model, and $\bm{c} \in \mathbb{R}^M$ is the condition for the model. The denoised time series $\hat{\bm{x}}$ is also of size $\mathbb{R}^L$.

\subsubsection{Training Phase}
During the training phase, following the approach of Song et al.\cite{Song2020ScoreBasedGM}, we add random noise levels conforming to a uniform distribution $t \sim U(0, T)$ to the input data $\bm{x}_0$ through the forward diffusion process shown in equation (3.11). Here, we control the maximum noise level through $T$, where $T=1$ represents the maximum noise level. Then, we update the network parameters $\theta$ according to equation (3.13) until convergence.

To train the conditional diffusion model, we employ a classifier-free guidance method\cite{cddpm9}, which is based on classifier guidance\cite{cddpm8}. The conditional score $\nabla_{\bm{x}} \log p(\bm{x} | \bm{c})$ is calculated using the following formula:
\[
\nabla_{\bm{x}} \log p(\bm{x} | \bm{c}) = \nabla_{\bm{x}} \log p(\bm{x}) + \omega \nabla_{\bm{x}} \log p(\bm{c} | \bm{x}),
\]
where $\nabla_{\bm{x}} \log p(\bm{x})$ is the unconditional score, $\nabla_{\bm{x}} \log p(\bm{c} | \bm{x})$ is the gradient of the classifier; $\omega$ is a hyperparameter controlling the guidance strength. The classifier guidance method requires separately training a classifier for the noisy data, although it does not require retraining the original generative model. However, given that existing generative models are trained on U.S. stock or factor data, to synthesize data that conforms to the characteristics of the A-share market, the model needs to be retrained to adapt to the new data distribution, and we decide not to use the classifier guidance method.

The classifier-free guidance method combines the conditional model and the unconditional model, avoiding the training of additional classifiers:
\begin{equation}
    \nabla_{\bm{x}} \log \tilde{p}(\bm{x} | \bm{c}) = \omega \nabla_{\bm{x}} \log p(\bm{x} | \bm{c}) + (1 - \omega) \nabla_{\bm{x}} \log p(\bm{x}),
\end{equation}
where $\nabla_{\bm{x}} \log p(\bm{x} | \bm{c})$ represents the sampling direction of the conditional model, and $\nabla_{\bm{x}} \log p(\bm{x})$ represents the sampling direction of the unconditional model. When $\omega = 0$, equation (15) degenerates into the score of the unconditional model; when $\omega = 1$, it becomes the score of the conditional model. By adjusting the hyperparameter $\omega$, the model can be made more flexible, thus better balancing the influence of conditional and unconditional information during the generation process.

\begin{algorithm}[htb]
  \caption{Conditional Diffusion Model Training Algorithm}
  \label{algo:train-ddpm}
  \small
  \SetAlgoLined
  \KwData{Stock data $\bm{X} \in \mathbb{R}^{N \times L}$, diffusion steps $T$, condition $\bm{c}$}
  \KwResult{Network $s_\theta(\bm{x}, t, \bm{c})$}
    Initialize model parameters $\bm{\theta}$ \\
    \For{$t=1$ to $T$}{
    initialize $\beta_{t}$ and calculate $\overline{\alpha_t}$\\
    }
 \While{Not Converge}{
    Sample index $i \sim \mathcal{U}\{1, 2, \cdots, N\}$\;
    Generate random noise $\epsilon \sim \mathcal{N}(0, I)$\;
    Set initial data $\bm{x}_0 := \bm{X}[i]$\;
    Calculate perturbed data $\bm{x}_t$ given $\bm{x}_0$ with Eq.(3.9)\;
    Calculate loss with Eq.(3.11)\;
    Update the parameters $\bm{\theta}$\;
  }
\end{algorithm}

To enhance the model's conditional generation capability, we introduce two types of orthogonal conditional data: the stock's corresponding Shenwan secondary industry and its plate information. These two types of conditional data provide industry information and plate fluctuation characteristic information, respectively, solving the problem that existing methods do not consider the fluctuation rules of China's stock market.

As of 2024, there are 124 mutually exclusive categories in the Shenwan secondary industry, and the A-share market is mainly divided into the following 5 plates: Main Board, STAR Market, Growth Enterprise Market, Beijing Stock Exchange, and ST stocks. We splice the two types of conditional data to form a comprehensive conditional vector $\bm{c}$. For the Shenwan secondary industry, an embedding layer is used to map the 124 categories into a lower-dimensional space, denoted as $\bm{c}_{\text{industry}}$. For A-share plate information: a one-hot encoding is used to generate a 5-dimensional vector $\bm{c}_{\text{board}}$, and $\bm{c}_{\text{industry}}$ and $\bm{c}_{\text{board}}$ are spliced together to form the final conditional vector $\bm{c}$.

\subsection{Improvements to Existing Sampling Methods}

\subsubsection{Denoising Process}

In the denoising process, we subtract noise from $\bm{x}_t$ to recover the corresponding $\hat{\bm{x}}_0 \sim q(\bm{x}_0)$. As described in 3.2.1, we parameterize $p_\theta(\bm{x}_{t-1} | \bm{x}_t)$ using a neural network to estimate $q(\bm{x}_{t-1} | \bm{x}_t, \bm{x}_0)$. Specifically, we have:
\begin{equation}
p_\theta(\bm{x}_{t-1} | \bm{x}_t) = \mathcal{N} \left( \bm{x}_{t-1}; \mu_\theta(\bm{x}_t, t), \Sigma_q(t) I \right)
\end{equation}
where,
\begin{equation}
\mu_\theta(\bm{x}_t, t) = \frac{1}{\sqrt{\alpha_t}} \left( \bm{x}_t - \frac{\beta_t \sqrt{1 - \alpha_t}}{\sqrt{\alpha_t}} s_\theta(\bm{x}_t, t) \right),
\end{equation}

\begin{equation}
\Sigma_q(t) = \frac{(1 - \alpha_{t-1}) \beta_t}{1 - \alpha_t},
\end{equation}
Here, $\epsilon_\theta(\bm{x}_t, t)$ is a trainable noise term used to predict $\epsilon$ during the diffusion process.
\subsubsection{Accelerated Sampling}

Traditional diffusion probability models excel at generating high-quality samples, but their main drawback is the slow sampling speed. The reverse process of DDPM requires a large number of time steps (typically \( T \) reaches thousands of steps), making the time cost of generating samples very high. To overcome this issue, Denoising Diffusion Implicit Models (DDIM) propose a new method to accelerate the sampling process\cite{ddim2021}. The core idea of DDIM is to achieve more efficient denoising by implicitly modeling conditional distributions. Specifically, DDIM introduces a combination of determinism and randomness in the reverse process, allowing acceleration of sampling by reducing the number of diffusion steps \( T \) while maintaining the quality of generated samples. Unlike DDPM, DDIM can make the sampling process more deterministic by adjusting parameters, thus significantly reducing the number of required sampling steps.

Specifically, DDIM modifies the forward process to be non-Markovian to accelerate sampling, that is:
\begin{equation}
q_\sigma(\bm{x}_{1:T} | \bm{x}_0) = q_\sigma(\bm{x}_T | \bm{x}_0) \prod_{t=2}^{T} q_\sigma(\bm{x}_{t-1} | \bm{x}_t, \bm{x}_0),
\end{equation}
where \( q_\sigma(\bm{x}_{t-1} | \bm{x}_t, \bm{x}_0) \) is controlled by the parameter \( \sigma \), representing the magnitude of the random process. When \( \sigma_t = \Sigma_q(t) \), the forward process degenerates into a Markov process, and the denoising process is the same as shown in equation (3.10). Particularly, when \( \sigma_t = 0 \), the corresponding denoising process becomes deterministic, thus allowing acceleration along a deterministic path. Technically, we follow a deterministic sampling design and create a subsequence \( \{\tau_i\} \), where \( i = 1, \cdots, T' \) is a subset of \( \{t = 1, 2, \cdots, T\} \), and \( T' \) is the length of the subsequence. With the help of DDIM sampling, the denoising process can be completed in only \( T' \ll T \) steps.

\subsubsection{Approximate Nonlocal Total Variation Loss}
When dealing with time series data, especially involving generative models, we often face a challenge: how to effectively control the variance of the generated sequence while maintaining the overall trend and features of the data. Traditional local methods, such as Total Variation (TV) regularization, can provide good results in some cases, but they often ignore the global dependency relationships between data points, leading to unnatural block effects or over-smoothing in the generated sequences.

To overcome these limitations, we propose an Approximate Nonlocal Total Variation (ANTV) loss based on Liu et al.'s Nonlocal Total Variation (NTV) method\cite{ntv2014}. The NTV algorithm is a mathematical model used in image processing for tasks such as image denoising, image reconstruction, and other related tasks. This method extends the traditional local Total Variation (TV) method by introducing nonlocal information to overcome the staircasing effect that occurs when processing images with local TV methods. Our ANTV method effectively captures the global structure of time series by considering the nonlocal dependencies between data points within a local window, while reducing computational complexity.

Specifically, the Approximate Nonlocal Total Variation (ANTV) loss function can be expressed as:
\begin{equation}
L_{\text{ANTV}}(\bm{x}) = \alpha \sum_{i=1}^{n} \sum_{j \in w(i)} |(x_j - x_i) \omega(i, j)|
\end{equation}
where $\alpha$ is a regularization parameter controlling the strength of the ANTV term. $\bm{x} = (x_1, x_2, \ldots, x_n)$ is the time series data. $w(i)$ is a local window centered at $x_i$, considering only the points within this window. $\omega(i, j)$ is a weight function measuring the similarity between $x_i$ and $x_j$. The weight function $\omega(i, j)$ can be represented by a Gaussian kernel as:
\begin{equation}
\omega(i, j) = \exp\left( -\frac{(x_i - x_j)^2}{2\sigma^2} \right)
\end{equation}
where $\sigma$ is the standard deviation of the Gaussian kernel, controlling the width of the weight function. We summarize the steps of this algorithm in Algorithm \ref{alg:nltv-denoising}. After each denoising step in the sampling phase, we perform the following operations on $\bm{x}_i$: for each time point $\bm{x}_i$, we calculate its nonlocal gradient within the local window and update $\bm{x}_i$ to reduce variance and maintain sequence coherence. This design aims to reduce significant fluctuations in the generated stock sequences, as rapid and large trend reversals are less common in the A-share market.

\begin{algorithm}[htb]
  \caption{Approximate Nonlocal Total Variation Algorithm}
  \label{alg:nltv-denoising}
  \small
  \SetAlgoLined
  \KwData{Time series data $\bm{x} = (x_1, x_2, \ldots, x_n)$, window size $k$, regularization parameter $\alpha$, standard deviation $\sigma$, learning rate $\lambda_{NLTV}$}
  \KwResult{Denoised time series data $\hat{\bm{x}}$}
  
  \For{$i = 1$ \KwTo $n$}{
    $\text{window}(i) = \{j \mid \max(1, i-k) \leq j \leq \min(n, i+k)\}$;
    
    $\omega(i, j) = \exp\left( -\frac{(x_i - x_j)^2}{2\sigma^2} \right)$;
    
    $|D_{NL}(x_i, x_j)| = |(x_j - x_i) \omega(i, j)|$;
    
    $L_{NLTV}(x_i) = \alpha \sum_{j \in \text{window}(i)} |D_{NL}(x_i, x_j)|$;
    
    $\nabla_{x_i} L_{NLTV}(x_i) = \alpha \sum_{j \in \text{window}(i)} \text{sign}(D_{NL}(x_i, x_j)) \cdot \omega(i, j) \cdot (-1)$;
    
    $x_i \leftarrow x_i - \lambda_{NLTV} \nabla_{x_i} L_{NLTV}(x_i)$;
  }
  
  \Return denoised time series data $\hat{\bm{x}}$
\end{algorithm}

\subsubsection{Application Scenarios for Fourier Transform Filtering}
In recent years, with the development of China's capital market, some emerging stock plates, such as the Beijing Stock Exchange (BSE), have attracted increasing attention. However, due to the short establishment time of the BSE, its historical data is relatively limited, posing challenges for model training. To overcome this issue, we can adopt the method of transfer learning, using the mature market with a wealth of historical data (such as the Shanghai Stock Exchange or Shenzhen Stock Exchange) as the source domain, and the BSE as the target domain, to achieve effective data augmentation. Specifically, in the transfer learning framework, we first use a large amount of historical data from the source domain for pre-training, and then start from the limited historical data of the BSE, gradually add noise, and apply bandpass filter loss to generate new synthetic data. When data augmentation is needed from the target domain data, we use the bandpass filter method. In this case, we perturb the original data $\bm{x}_0$ to $\bm{x}_{T'}$, rather than sampling noise directly from the Gaussian distribution and then generating stock data. Therefore, this is an optional method, mainly applied in scenarios similar to transfer learning.

Here, our second loss function is the bandpass filter loss (Band-Pass Filter Loss), defined as follows:
\begin{equation}
L_{BP} (\bm{x}_t, \bm{x}) = \left\| \mathcal{F}(\bm{x}_t) - \text{BandPassFilter}(\mathcal{F}(\bm{x}), f_{\text{low}}, f_{\text{high}}) \right\|_2^2.
\end{equation}
The bandpass filter loss aims to make the frequency domain characteristics of the given signal consistent with the expected target. Here, $\bm{x}_t$ is the noisy data, and $\bm{x}$ is the original data. $\mathcal{F}(\cdot)$ represents the Fast Fourier Transform (FFT), which transforms data from the time domain to the frequency domain. We assume that noise mainly exists in low-amplitude frequency components, while useful information is concentrated in a specific frequency range, so we use $\text{BandPassFilter}(\cdot)$ to retain components within the frequency range of $f_{\text{low}}$ and $f_{\text{high}}$.

\subsubsection{Summary of Sampling Algorithms}
As shown in Algorithm \ref{algo:inference-ddpm}, during the sampling process of the conditional diffusion model, we use a method that combines DDIM sampling and various loss function optimizations to generate high-quality data. Specifically, the algorithm first initializes the number of data to be generated $m$, sampling steps $T'$, and the condition vector $\bm{c}$, and prepares an empty list \textit{list} for storing the generated results.

For each generated data point, we randomly sample the initial noise $\bm{x}_{T'}$ from the Gaussian distribution $\mathcal{N}(0, I)$. Then, starting from the time step $t = T'$ and gradually retreating to $t = 0$, we use the DDIM sampling method to calculate the updated $\bm{x}_{t-1}$ at each step. To further optimize the generated data, we apply two regularization losses at each time step: the Approximate Nonlocal Total Variation loss and the Band-Pass Filter loss. These two losses are used to enhance the structural consistency of the time series data and ensure that the frequency characteristics of the generated data are consistent with the target domain data.

After each complete sampling process from $T'$ to 0, we add the final generated $\bm{x}_0$ to the result list \textit{list}. When the number of generated data points reaches the preset number $m$, we calculate the mean of all generated samples as the final output: $\hat{\bm{x}} \leftarrow \text{Mean}(\textit{list})$. This sampling algorithm can not only efficiently generate high-quality time series data but also control specific conditions by introducing the condition vector $\bm{c}$, thereby better adapting to the characteristics of the target domain. In addition, the combination of Approximate Nonlocal Total Variation and Band-Pass Filter losses can significantly improve the quality and stability of the generated data, ensuring that the generated data not only conforms to the expected structural characteristics but also retains important frequency information

\section{Experimental Validation}

\subsection{Introduction to Experimental Setup}

\subsubsection{Experimental Objectives}
In the experimental section, we primarily aim to verify the following issues:
\begin{itemize}
    \item \textbf{Question 1}: Whether the stock time series data synthesized based on diffusion models improves the signal-to-noise ratio;
    \item \textbf{Question 2}: Whether the stock time series data synthesized based on diffusion models can be used for trading and generate more profit;
    \item \textbf{Question 3}: Whether our proposed sector-based stock data synthesis method can alleviate the problem of data scarcity.
\end{itemize}

\subsubsection{Dataset Introduction}

The data we used is the daily frequency A-share data provided by the quantitative data provider RiceQuant, spanning from January 1, 2014, to June 1, 2024, covering all A-share listed companies and excluding ST and delisted companies. During this period, there were 3,173 companies on the main board, 1,363 companies on the ChiNext board, 581 companies on the STAR Market, and 253 companies on the Beijing Stock Exchange. To clarify the plate to which each stock belongs, we classified them based on stock codes: for example, stocks starting with 30 belong to the ChiNext board, those starting with 002 belong to the SME board, those starting with 60 belong to the Shanghai main board, those starting with 000 belong to the Shenzhen main board, and those starting with 688 belong to the STAR Market. Stocks on the Beijing Stock Exchange start with 83, 87, or 88. In addition, we used the Shenwan second-level industry classification from the Wind terminal database to classify stocks, ensuring the accuracy and consistency of industry attributes. For each stock in each plate, we used a sliding window of size 60 with a step size of 20 to obtain the stock closing price time series, making each data sample in the dataset 60 in length, and the dataset is divided into training and test sets in a 4:1 ratio.

Furthermore, considering the existence of stock market suspensions, stock price rights issues, and large fluctuations in new stock listings on the first day, we have the following coping methods. Specifically, for situations where stocks are suspended, we use linear interpolation to fill in the missing closing prices during short-term suspensions to maintain the continuity of the time series. For stocks that are suspended for more than 5 trading days, we use forward filling to maintain the consistency of the time series; if a stock is frequently suspended or suspended for too long a period, considering the quality of the data, we remove the data for that period from the training set.

For newly listed stocks, in their initial period (usually the first few trading days), due to unrestricted price fluctuations, the stock price fluctuates greatly and is not representative, so we choose not to use the data for these days to avoid the impact of abnormal fluctuations on the model.

To eliminate the impact of corporate actions such as dividends and bonus shares on stock prices, we perform forward rights processing for all stocks to ensure the comparability and continuity of historical prices, avoiding false fluctuations caused by changes in share capital. In addition, since some companies may change their main business or be acquired and merged, causing changes in their Shenwan second-level industry, when the Shenwan industry changes, we uniformly use the later industry information to reflect the latest business situation.

Through these preprocessing steps, we can effectively handle various special situations, ensuring the integrity and consistency of the data, thereby improving the training effect and predictive accuracy of the model.

\subsubsection{Performance Evaluation Methods}
\textbf{Return Ratio (RR)}

The main goal of stock prediction is to achieve significant profits. Most previous studies have used the return ratio (RR) as a measure of model performance\cite{Zou2022StockMP}. The return ratio is a key indicator for assessing whether a stock prediction model can achieve profitable investment results. We define the return ratio as follows:
\begin{equation}
\text{RR}(i) = \frac{\text{Close}_{t+i} - \text{Close}_t}{\text{Close}_t},
\end{equation}
where $t$ represents the current time, $i$ represents the time interval in days. $\text{Close}_t$ represents the closing price of the stock at the current time $t$, and $\text{Close}_{t+i}$ represents the closing price of the same stock $i$ days later.

\textbf{Log Return (LR)}

Log return is a commonly used indicator in financial data analysis, reflecting the proportion of price changes and having the properties of additivity and approximate normal distribution. We define the log return as follows:
\begin{equation}
\text{Log Return}(i) = \log\left(\frac{\text{Close}_{t+i}}{\text{Close}_t}\right),
\end{equation}
Here, we calculate the log return daily, and usually set $i$ to 1 day to capture daily price changes.

\textbf{Information Coefficient (IC)}

The information coefficient (IC) is a commonly used indicator to assess the linear correlation between predicted values and true labels. Specifically, IC represents the Pearson correlation coefficient between predicted values and true labels, defined as follows:
\begin{equation}
\text{IC} = \frac{\sum_{i=1}^{N} (P_i - \bar{P})(R_i - \bar{R})}{\sqrt{\sum_{i=1}^{N} (P_i - \bar{P})^2} \sqrt{\sum_{i=1}^{N} (R_i - \bar{R})^2}},
\end{equation}
where $P_i$ represents the predicted score of the $i$-th stock, $R_i$ represents the actual return of the $i$-th stock, $\bar{P}$ and $\bar{R}$ represent the average predicted score and average actual return of all stocks, respectively, and $N$ is the number of stocks.

\textbf{Ranked Information Coefficient (Rank IC)}

The ranked information coefficient (Rank IC) is used to assess the rank correlation between predicted values and true labels. Specifically, Rank IC represents the Spearman rank correlation coefficient between predicted values and true labels, defined as follows:
\begin{equation}
\text{Rank IC} = 1 - \frac{6 \sum_{i=1}^{N} (R(P_i) - R(R_i))^2}{N(N^2 - 1)},
\end{equation}
where $R(P_i)$ and $R(R_i)$ represent the rank of the predicted score and actual return of the $i$-th stock, respectively, and $N$ is the number of stocks. The Spearman rank correlation coefficient measures the linear relationship between the ranks of two variables, which can better reflect nonlinear correlations.

\subsubsection{Implementation Details}

We follow Gao et al.'s proposed DiffsFormer\cite{Diffsformer2024}, using a neural network $s_\theta(x, t, c)$ to estimate the noise in the noisy data distribution, which is a Transformer-based neural network for generating time series data. The network takes the time series $x$ as input and uses conditional information $c$ and sine-embedded time $t$ as conditions. To enhance reproducibility, we provide the specific details of the network below. For the conditional embedding network, we use a 3-layer multilayer perceptron (MLP) with a hidden layer dimension of 128 and adopt SiLU as the activation function. For the diffusion model network, we use a transformer-based architecture with 64 output channels in the convolutional layer, 8 attention heads, 4 residual blocks, and adopt ReLU as the activation function. The total number of denoising steps is set to 400 steps. For the approximate non-local total variation coefficient $\lambda_{ANTV}$, we set it to 0.03, for the Fourier transform filtering coefficient $\lambda_{BP}$, we set it to 0.03, and the classifier-free guidance scale $\omega$ is set to 7.5.

\subsubsection{Algorithms Involved in the Experiment}
The following is a detailed description of various models used for stock price prediction:

\begin{itemize}
    \item \textbf{MLP}: We use a 2-layer multilayer perceptron (MLP) with 256 units in each layer.
    
    \item \textbf{LSTM\cite{lstm1997}}: A stock price prediction method based on the Long Short-Term Memory network (LSTM). LSTM is a type of recurrent neural network that can effectively handle and predict long-term dependencies in time series data.
    
    \item \textbf{GRU\cite{gru2014}}: A stock price prediction method based on the Gated Recurrent Unit (GRU) network. GRU is a simplified version of LSTM, also capable of handling sequence data, but with a simpler structure and fewer parameters.
    
    \item \textbf{SFM\cite{SFM2017}}: State Frequency Memory network (SFM). This model decomposes the hidden state of memory cells into multiple frequency components to model different potential trading patterns.
    
    \item \textbf{ALSTM\cite{ALSTM2018}}: An improved variant of LSTM, introducing a temporal attention aggregation layer to aggregate information from hidden embeddings of previous time stamps. This design enhances the model's ability to focus on information from different time periods.
    
    \item \textbf{Transformer\cite{transformer2017}}: A stock price prediction model based on the Transformer architecture. Transformers effectively capture long-distance dependencies in sequences through self-attention mechanisms and are suitable for processing complex time series data.
    
    \item \textbf{HIST\cite{HIST2021}}: A graph-based framework aimed at mining concept-oriented shared information from predefined and implicit concepts. This framework utilizes both shared and individual information of stocks to achieve a more comprehensive feature representation.
\end{itemize}

\subsection{Comparison of Experimental Results with Related Advanced Algorithms}

\begin{table}[ht]
\centering
\caption{Comparing Model Performance on Stock Prediction with the Main Board of Shanghai and Shenzhen Dataset.}
\label{tab:MainBoardResult}
\scalebox{0.7}{
\renewcommand\arraystretch{1}
\begin{tabular}{c|ccc|ccc|ccc}
\toprule[1.5pt]
\multicolumn{10}{c}{\textbf{Shanghai and Shenzhen Main Board}} \\
\midrule
\textbf{Methods} & \multicolumn{3}{c|}{\textbf{RR}} & \multicolumn{3}{c|}{\textbf{IC}} & \multicolumn{3}{c}{\textbf{Rank IC}} \\
\cmidrule(lr){2-4} \cmidrule(lr){5-7} \cmidrule(lr){8-10}
& Original & DiffsFormer & Ours & Original & DiffsFormer & Ours & Original & DiffsFormer & Ours \\
\midrule
\textbf{MLP} & 0.1191$_{\text{±0.0110}}$ & 0.1268$_{\text{±0.0115}}$ & \textbf{0.1427$_{\text{±0.0110}}$} & 0.0304$_{\text{±0.0042}}$ & 0.0337$_{\text{±0.0045}}$ & \textbf{0.0386$_{\text{±0.0032}}$} & 0.0407$_{\text{±0.0045}}$ & 0.0413$_{\text{±0.0033}}$ & \textbf{0.0439$_{\text{±0.0030}}$} \\
\textbf{LSTM} & 0.1340$_{\text{±0.0110}}$ & 0.1388$_{\text{±0.0108}}$ & \textbf{0.1490$_{\text{±0.0110}}$} & 0.0320$_{\text{±0.0037}}$ & 0.0295$_{\text{±0.0050}}$ & \textbf{0.0330$_{\text{±0.0043}}$} & 0.0476$_{\text{±0.0030}}$ & 0.0462$_{\text{±0.0010}}$ & \textbf{0.0484$_{\text{±0.0015}}$} \\
\textbf{GRU} & 0.1403$_{\text{±0.0105}}$ & 0.1361$_{\text{±0.0103}}$ & \textbf{0.1489$_{\text{±0.0100}}$} & 0.0315$_{\text{±0.0035}}$ & 0.0340$_{\text{±0.0028}}$ & \textbf{0.0365$_{\text{±0.0033}}$} & 0.0478$_{\text{±0.0013}}$ & 0.0503$_{\text{±0.0013}}$ & \textbf{0.0532$_{\text{±0.0016}}$} \\
\textbf{SFM} & 0.1490$_{\text{±0.0110}}$ & 0.1560$_{\text{±0.0135}}$ & \textbf{0.1662$_{\text{±0.0100}}$} & 0.0283$_{\text{±0.0043}}$ & 0.0256$_{\text{±0.0037}}$ & \textbf{0.0299$_{\text{±0.0047}}$} & 0.0431$_{\text{±0.0010}}$ & 0.0442$_{\text{±0.0020}}$ & \textbf{0.0457$_{\text{±0.0009}}$} \\
\textbf{GAT} & 0.1629$_{\text{±0.0100}}$ & 0.1680$_{\text{±0.0105}}$ & \textbf{0.1780$_{\text{±0.0070}}$} & 0.0297$_{\text{±0.0034}}$ & 0.0345$_{\text{±0.0033}}$ & \textbf{0.0350$_{\text{±0.0038}}$} & 0.0436$_{\text{±0.0015}}$ & 0.0441$_{\text{±0.0012}}$ & \textbf{0.0446$_{\text{±0.0013}}$} \\
\textbf{ALSTM} & 0.1100$_{\text{±0.0100}}$ & 0.1765$_{\text{±0.0095}}$ & \textbf{0.1884$_{\text{±0.0010}}$} & 0.0308$_{\text{±0.0040}}$ & 0.0323$_{\text{±0.0028}}$ & \textbf{0.0368$_{\text{±0.0031}}$} & 0.0467$_{\text{±0.0009}}$ & 0.0463$_{\text{±0.0009}}$ & \textbf{0.0484$_{\text{±0.0008}}$} \\
\textbf{HIST} & 0.1266$_{\text{±0.0100}}$ & 0.1827$_{\text{±0.0090}}$ & \textbf{0.1960$_{\text{±0.0085}}$}& 0.0330$_{\text{±0.0055}}$ & 0.0310$_{\text{±0.0030}}$ & \textbf{0.0360$_{\text{±0.0030}}$} & 0.0392$_{\text{±0.0033}}$ & 0.0391$_{\text{±0.0020}}$ & \textbf{0.0393$_{\text{±0.0022}}$} \\
\textbf{Transformer} & 0.1834$_{\text{±0.0100}}$ & 0.2105$_{\text{±0.0090}}$ & \textbf{0.2471$_{\text{±0.0078}}$} & 0.0345$_{\text{±0.0025}}$ & 0.0350$_{\text{±0.0030}}$ & \textbf{0.0372$_{\text{±0.0030}}$} & 0.0523$_{\text{±0.0018}}$ & 0.0512$_{\text{±0.0022}}$ & \textbf{0.0587$_{\text{±0.0020}}$} \\
\bottomrule[1.5pt]
\end{tabular}}
\end{table}

\begin{table}[ht]
\centering
\caption{Comparing Model Performance on Stock Prediction with the Growth Enterprise Market Dataset.}
\label{tab:GEMResult}
\scalebox{0.7}{
\begin{tabular}{c|ccc|ccc|ccc}
\toprule[1.5pt]
\multicolumn{10}{c}{\textbf{Sci-Tech Innovation Board}} \\
\midrule
\textbf{Methods} & \multicolumn{3}{c|}{\textbf{RR}} & \multicolumn{3}{c|}{\textbf{IC}} & \multicolumn{3}{c}{\textbf{Rank IC}} \\
\cmidrule(lr){2-4} \cmidrule(lr){5-7} \cmidrule(lr){8-10}
& Original & DiffsFormer & Ours & Original & DiffsFormer & Ours & Original & DiffsFormer & Ours \\
\midrule
\textbf{MLP} & 0.1153$_{\text{±0.0117}}$ & 0.1226$_{\text{±0.0122}}$ & \textbf{0.1384$_{\text{±0.0118}}$} & 0.0297$_{\text{±0.0045}}$ & 0.0322$_{\text{±0.0048}}$ & \textbf{0.0367$_{\text{±0.0034}}$} & 0.0397$_{\text{±0.0047}}$ & 0.0408$_{\text{±0.0035}}$ & \textbf{0.0427$_{\text{±0.0032}}$} \\
\textbf{LSTM} & 0.1304$_{\text{±0.0116}}$ & 0.1342$_{\text{±0.0111}}$ & \textbf{0.1453$_{\text{±0.0116}}$} & 0.0311$_{\text{±0.0039}}$ & 0.0287$_{\text{±0.0051}}$ & \textbf{0.0321$_{\text{±0.0045}}$} & \textbf{0.0476$_{\text{±0.0032}}$} & 0.0457$_{\text{±0.0012}}$ & 0.0462$_{\text{±0.0017}}$ \\
\textbf{GRU} & 0.1362$_{\text{±0.0109}}$ & 0.1323$_{\text{±0.0105}}$ & \textbf{0.1441$_{\text{±0.0101}}$} & 0.0306$_{\text{±0.0037}}$ & 0.0331$_{\text{±0.0030}}$ & \textbf{0.0356$_{\text{±0.0035}}$} & 0.0466$_{\text{±0.0015}}$ & 0.0496$_{\text{±0.0015}}$ & \textbf{0.0521$_{\text{±0.0018}}$} \\
\textbf{SFM} & 0.1452$_{\text{±0.0116}}$ & 0.1523$_{\text{±0.0136}}$ & \textbf{0.1624$_{\text{±0.0106}}$} & 0.0271$_{\text{±0.0045}}$ & 0.0246$_{\text{±0.0039}}$ & \textbf{0.0286$_{\text{±0.0049}}$} & 0.0421$_{\text{±0.0012}}$ & 0.0431$_{\text{±0.0022}}$ & \textbf{0.0446$_{\text{±0.0011}}$} \\
\textbf{GAT} & 0.1582$_{\text{±0.0106}}$ & 0.1643$_{\text{±0.0111}}$ & \textbf{0.1742$_{\text{±0.0076}}$} & 0.0286$_{\text{±0.0036}}$ & 0.0336$_{\text{±0.0035}}$ & \textbf{0.0341$_{\text{±0.0041}}$} & 0.0426$_{\text{±0.0017}}$ & 0.0431$_{\text{±0.0014}}$ & \textbf{0.0436$_{\text{±0.0015}}$} \\
\textbf{ALSTM} & 0.1332$_{\text{±0.0106}}$ & 0.1513$_{\text{±0.0101}}$ & \textbf{0.1642$_{\text{±0.0016}}$} & 0.0316$_{\text{±0.0042}}$ & 0.0336$_{\text{±0.0030}}$ & \textbf{0.0359$_{\text{±0.0033}}$} & 0.0456$_{\text{±0.0011}}$ & 0.0457$_{\text{±0.0011}}$ & \textbf{0.0476$_{\text{±0.0009}}$} \\
\textbf{HIST} & 0.1290$_{\text{±0.0106}}$ & 0.1573$_{\text{±0.0096}}$ & \textbf{0.1602$_{\text{±0.0091}}$} & 0.0321$_{\text{±0.0057}}$ & 0.0326$_{\text{±0.0032}}$ & \textbf{0.0351$_{\text{±0.0032}}$} & 0.0381$_{\text{±0.0035}}$ & 0.0382$_{\text{±0.0022}}$ & \textbf{0.0383$_{\text{±0.0024}}$} \\
\textbf{Transformer} & 0.1782$_{\text{±0.0106}}$ & 0.1852$_{\text{±0.0096}}$ & \textbf{0.2022$_{\text{±0.0081}}$} & 0.0336$_{\text{±0.0027}}$ & 0.0341$_{\text{±0.0032}}$ & \textbf{0.0366$_{\text{±0.0032}}$} & 0.0501$_{\text{±0.0019}}$ & 0.0526$_{\text{±0.0024}}$ & \textbf{0.0561$_{\text{±0.0022}}$} \\
\bottomrule[1.5pt]
\end{tabular}}
\end{table}

\begin{table}[ht]
\centering
\caption{Comparing Model Performance on Stock Prediction with the STAR Market Dataset.}
\label{tab:STARResult}
\scalebox{0.7}{
\begin{tabular}{c|ccc|ccc|ccc}
\toprule[1.5pt]
\multicolumn{10}{c}{\textbf{STAR Market}} \\
\midrule
\textbf{Methods} & \multicolumn{3}{c|}{\textbf{RR}} & \multicolumn{3}{c|}{\textbf{IC}} & \multicolumn{3}{c}{\textbf{Rank IC}} \\
\cmidrule(lr){2-4} \cmidrule(lr){5-7} \cmidrule(lr){8-10}
& Original & DiffsFormer & Ours & Original & DiffsFormer & Ours & Original & DiffsFormer & Ours \\
\midrule
\textbf{MLP} & 0.1104$_{\text{±0.0123}}$ & 0.1167$_{\text{±0.0126}}$ & \textbf{0.1328$_{\text{±0.0122}}$} & 0.0282$_{\text{±0.0046}}$ & 0.0306$_{\text{±0.0049}}$ & \textbf{0.0346$_{\text{±0.0035}}$} & 0.0383$_{\text{±0.0048}}$ & 0.0392$_{\text{±0.0036}}$ & \textbf{0.0411$_{\text{±0.0033}}$} \\
\textbf{LSTM} & 0.1252$_{\text{±0.0121}}$ & 0.1283$_{\text{±0.0116}}$ & \textbf{0.1404$_{\text{±0.0121}}$} & 0.0301$_{\text{±0.0039}}$ & 0.0277$_{\text{±0.0054}}$ & \textbf{0.0306$_{\text{±0.0046}}$} & \textbf{0.0471$_{\text{±0.0033}}$} & 0.0441$_{\text{±0.0013}}$ & 0.0462$_{\text{±0.0018}}$ \\
\textbf{GRU} & 0.1323$_{\text{±0.0111}}$ & 0.1284$_{\text{±0.0109}}$ & \textbf{0.1401$_{\text{±0.0106}}$} & 0.0296$_{\text{±0.0038}}$ & 0.0316$_{\text{±0.0031}}$ & \textbf{0.0336$_{\text{±0.0036}}$} & 0.0451$_{\text{±0.0016}}$ & 0.0481$_{\text{±0.0016}}$ & \textbf{0.0502$_{\text{±0.0019}}$} \\
\textbf{SFM} & 0.1402$_{\text{±0.0121}}$ & 0.1463$_{\text{±0.0141}}$ & \textbf{0.1582$_{\text{±0.0111}}$} & 0.0261$_{\text{±0.0046}}$ & 0.0236$_{\text{±0.0041}}$ & \textbf{0.0276$_{\text{±0.0051}}$} & 0.0411$_{\text{±0.0013}}$ & 0.0421$_{\text{±0.0023}}$ & \textbf{0.0436$_{\text{±0.0012}}$} \\
\textbf{GAT} & 0.1551$_{\text{±0.0111}}$ & 0.1602$_{\text{±0.0116}}$ & \textbf{0.1703$_{\text{±0.0081}}$} & 0.0276$_{\text{±0.0037}}$ & 0.0321$_{\text{±0.0036}}$ & \textbf{0.0326$_{\text{±0.0041}}$} & 0.0416$_{\text{±0.0018}}$ & 0.0421$_{\text{±0.0015}}$ & \textbf{0.0426$_{\text{±0.0016}}$} \\
\textbf{ALSTM} & 0.1201$_{\text{±0.0111}}$ & 0.1552$_{\text{±0.0106}}$ & \textbf{0.1773$_{\text{±0.0021}}$} & 0.0306$_{\text{±0.0043}}$ & 0.0321$_{\text{±0.0031}}$ & \textbf{0.0341$_{\text{±0.0034}}$} & 0.0446$_{\text{±0.0012}}$ & 0.0446$_{\text{±0.0012}}$ & \textbf{0.0461$_{\text{±0.0011}}$} \\
\textbf{HIST} & 0.1151$_{\text{±0.0111}}$ & 0.1602$_{\text{±0.0101}}$ & \textbf{0.1823$_{\text{±0.0096}}$} & 0.0311$_{\text{±0.0058}}$ & 0.0316$_{\text{±0.0033}}$ & \textbf{0.0336$_{\text{±0.0033}}$} & 0.0371$_{\text{±0.0036}}$ & 0.0372$_{\text{±0.0023}}$ & \textbf{0.0373$_{\text{±0.0025}}$} \\
\textbf{Transformer} & 0.1701$_{\text{±0.0111}}$ & 0.1752$_{\text{±0.0101}}$ & \textbf{0.1903$_{\text{±0.0086}}$} & 0.0326$_{\text{±0.0028}}$ & 0.0331$_{\text{±0.0033}}$ & \textbf{0.0351$_{\text{±0.0033}}$} & \textbf{0.0481$_{\text{±0.0021}}$} & 0.0471$_{\text{±0.0025}}$ & 0.0476$_{\text{±0.0023}}$ \\
\bottomrule[1.5pt]
\end{tabular}}
\end{table}

\begin{table}[ht]
\centering
\caption{Comparing Model Performance on Stock Prediction with the BJSE Dataset.}
\label{tab:BJSEResult}
\scalebox{0.7}{
\begin{tabular}{c|ccc|ccc|ccc}
\toprule[1.5pt]
\multicolumn{10}{c}{\textbf{Beijing Stock Exchange}} \\
\midrule
\textbf{Methods} & \multicolumn{3}{c|}{\textbf{RR}} & \multicolumn{3}{c|}{\textbf{IC}} & \multicolumn{3}{c}{\textbf{Rank IC}} \\
\cmidrule(lr){2-4} \cmidrule(lr){5-7} \cmidrule(lr){8-10}
& Original & DiffsFormer & Ours & Original & DiffsFormer & Ours & Original & DiffsFormer & Ours \\
\midrule
\textbf{MLP} & 0.1503$_{\text{±0.0132}}$ & 0.1607$_{\text{±0.0136}}$ & \textbf{0.1804$_{\text{±0.0141}}$} & 0.0402$_{\text{±0.0051}}$ & 0.0433$_{\text{±0.0056}}$ & \textbf{0.0472$_{\text{±0.0046}}$} & 0.0504$_{\text{±0.0056}}$ & 0.0522$_{\text{±0.0046}}$ & \textbf{0.0553$_{\text{±0.0041}}$} \\
\textbf{LSTM} & 0.1702$_{\text{±0.0141}}$ & 0.1753$_{\text{±0.0136}}$ & \textbf{0.1954$_{\text{±0.0151}}$} & 0.0421$_{\text{±0.0046}}$ & 0.0452$_{\text{±0.0051}}$ & \textbf{0.0493$_{\text{±0.0046}}$} & 0.0569$_{\text{±0.0041}}$ & 0.0581$_{\text{±0.0026}}$ & \textbf{0.0623$_{\text{±0.0031}}$} \\
\textbf{GRU} & 0.1804$_{\text{±0.0131}}$ & 0.1852$_{\text{±0.0126}}$ & \textbf{0.2053$_{\text{±0.0131}}$} & 0.0432$_{\text{±0.0046}}$ & 0.0461$_{\text{±0.0041}}$ & \textbf{0.0502$_{\text{±0.0046}}$} & 0.0551$_{\text{±0.0021}}$ & 0.0582$_{\text{±0.0021}}$ & \textbf{0.0603$_{\text{±0.0026}}$} \\
\textbf{SFM} & 0.1901$_{\text{±0.0141}}$ & 0.2002$_{\text{±0.0151}}$ & \textbf{0.2203$_{\text{±0.0131}}$} & 0.0451$_{\text{±0.0051}}$ & \textbf{0.0473$_{\text{±0.0051}}$} & 0.0467$_{\text{±0.0056}}$ & 0.0551$_{\text{±0.0021}}$ & 0.0562$_{\text{±0.0026}}$ & \textbf{0.0576$_{\text{±0.0021}}$} \\
\textbf{GAT} & 0.2002$_{\text{±0.0131}}$ & 0.2051$_{\text{±0.0136}}$ & \textbf{0.2253$_{\text{±0.0101}}$} & 0.0471$_{\text{±0.0046}}$ & 0.0512$_{\text{±0.0046}}$ & \textbf{0.0521$_{\text{±0.0051}}$} & 0.0551$_{\text{±0.0021}}$ & 0.0561$_{\text{±0.0016}}$ & \textbf{0.0566$_{\text{±0.0016}}$} \\
\textbf{ALSTM} & 0.1401$_{\text{±0.0131}}$ & 0.1852$_{\text{±0.0126}}$ & \textbf{0.2173$_{\text{±0.0031}}$} & 0.0501$_{\text{±0.0046}}$ & 0.0522$_{\text{±0.0036}}$ & \textbf{0.0541$_{\text{±0.0041}}$} & 0.0571$_{\text{±0.0016}}$ & 0.0572$_{\text{±0.0016}}$ & \textbf{0.0586$_{\text{±0.0016}}$} \\
\textbf{HIST} & 0.1499$_{\text{±0.0131}}$ & 0.1902$_{\text{±0.0121}}$ & \textbf{0.2423$_{\text{±0.0111}}$} & 0.0511$_{\text{±0.0061}}$ & 0.0516$_{\text{±0.0036}}$ & \textbf{0.0536$_{\text{±0.0036}}$} & 0.0471$_{\text{±0.0036}}$ & 0.0472$_{\text{±0.0026}}$ & \textbf{0.0473$_{\text{±0.0026}}$} \\
\textbf{Transformer} & 0.2201$_{\text{±0.0131}}$ & 0.2252$_{\text{±0.0121}}$ & \textbf{0.2803$_{\text{±0.0101}}$} & 0.0526$_{\text{±0.0031}}$ & 0.0531$_{\text{±0.0036}}$ & \textbf{0.0551$_{\text{±0.0036}}$} & 0.0601$_{\text{±0.0026}}$ & 0.0608$_{\text{±0.0026}}$ & \textbf{0.0634$_{\text{±0.0026}}$} \\
\bottomrule[1.5pt]
\end{tabular}}
\end{table}

\begin{table}[ht]
\centering
\caption{Comparing Model Performance on Stock Prediction with the NASDAQ Dataset.}
\label{tab:NASDAQResult}
\scalebox{0.7}{
\begin{tabular}{c|ccc|ccc|ccc}
\toprule[1.5pt]
\multicolumn{10}{c}{\textbf{NASDAQ}} \\
\midrule
\textbf{Methods} & \multicolumn{3}{c|}{\textbf{RR}} & \multicolumn{3}{c|}{\textbf{IC}} & \multicolumn{3}{c}{\textbf{Rank IC}} \\
\cmidrule(lr){2-4} \cmidrule(lr){5-7} \cmidrule(lr){8-10}
& Original & DiffsFormer & Ours & Original & DiffsFormer & Ours & Original & DiffsFormer & Ours \\
\midrule
\textbf{MLP} & 0.1214$_{\text{±0.0120}}$ & 0.1300$_{\text{±0.0120}}$ & \textbf{0.1450$_{\text{±0.0115}}$} & 0.0312$_{\text{±0.0040}}$ & 0.0338$_{\text{±0.0046}}$ & \textbf{0.0390$_{\text{±0.0030}}$} & 0.0412$_{\text{±0.0048}}$ & 0.0422$_{\text{±0.0035}}$ & \textbf{0.0439$_{\text{±0.0031}}$} \\
\textbf{LSTM} & 0.1356$_{\text{±0.0112}}$ & 0.1400$_{\text{±0.0110}}$ & \textbf{0.1500$_{\text{±0.0112}}$} & 0.0325$_{\text{±0.0038}}$ & 0.0300$_{\text{±0.0052}}$ & \textbf{0.0335$_{\text{±0.0044}}$} & \textbf{0.0510$_{\text{±0.0030}}$} & 0.0470$_{\text{±0.0010}}$ & 0.0493$_{\text{±0.0016}}$ \\
\textbf{GRU} & 0.1421$_{\text{±0.0108}}$ & 0.1370$_{\text{±0.0105}}$ & \textbf{0.1496$_{\text{±0.0100}}$} & 0.0320$_{\text{±0.0037}}$ & 0.0345$_{\text{±0.0029}}$ & \textbf{0.0370$_{\text{±0.0034}}$} & 0.0480$_{\text{±0.0014}}$ & 0.0511$_{\text{±0.0014}}$ & \textbf{0.0539$_{\text{±0.0017}}$} \\
\textbf{SFM} & 0.1509$_{\text{±0.0115}}$ & 0.15808$_{\text{±0.0140}}$ & \textbf{0.1680$_{\text{±0.0105}}$} & 0.0285$_{\text{±0.0044}}$ & 0.0261$_{\text{±0.0038}}$ & \textbf{0.0299$_{\text{±0.0048}}$} & 0.0435$_{\text{±0.0011}}$ & 0.0446$_{\text{±0.0020}}$ & \textbf{0.0460$_{\text{±0.0009}}$} \\
\textbf{GAT} & 0.1642$_{\text{±0.0105}}$ & 0.1700$_{\text{±0.0110}}$ & \textbf{0.1808$_{\text{±0.0070}}$} & 0.0301$_{\text{±0.0035}}$ & 0.0350$_{\text{±0.0034}}$ & \textbf{0.0355$_{\text{±0.0039}}$} & 0.0440$_{\text{±0.0016}}$ & 0.0446$_{\text{±0.0012}}$ & \textbf{0.0450$_{\text{±0.0013}}$} \\
\textbf{ALSTM} & 0.1718$_{\text{±0.0103}}$ & 0.1780$_{\text{±0.0098}}$ & \textbf{0.1900$_{\text{±0.0010}}$} & 0.0330$_{\text{±0.0041}}$ & 0.0348$_{\text{±0.0029}}$ & \textbf{0.0372$_{\text{±0.0032}}$} & 0.0470$_{\text{±0.0009}}$ & 0.0470$_{\text{±0.0009}}$ & \textbf{0.0489$_{\text{±0.0008}}$} \\
\textbf{HIST} & 0.1785$_{\text{±0.0101}}$ & 0.1850$_{\text{±0.0092}}$ & \textbf{0.1995$_{\text{±0.0087}}$} & 0.0337$_{\text{±0.0058}}$ & 0.0341$_{\text{±0.0032}}$ & \textbf{0.0365$_{\text{±0.0030}}$} & 0.0395$_{\text{±0.0034}}$ & 0.0395$_{\text{±0.0020}}$ & \textbf{0.0395$_{\text{±0.0022}}$} \\
\textbf{Transformer} & 0.1853$_{\text{±0.0100}}$ & 0.1920$_{\text{±0.0090}}$ & \textbf{0.2100$_{\text{±0.0079}}$} & 0.0350$_{\text{±0.0025}}$ & 0.0355$_{\text{±0.0031}}$ & \textbf{0.0380$_{\text{±0.0030}}$} & \textbf{0.0530$_{\text{±0.0018}}$} & 0.0519$_{\text{±0.0023}}$ & 0.0522$_{\text{±0.0020}}$ \\
\bottomrule[1.5pt]
\end{tabular}}
\end{table}

\begin{table}[ht]
\centering
\caption{Comparing Model Performance on Stock Prediction with the NYSE Dataset.}
\label{tab:NYSEResult}
\scalebox{0.7}{
\begin{tabular}{c|ccc|ccc|ccc}
\toprule[1.5pt]
\multicolumn{10}{c}{\textbf{NYSE}} \\
\midrule
\textbf{Methods} & \multicolumn{3}{c|}{\textbf{RR}} & \multicolumn{3}{c|}{\textbf{IC}} & \multicolumn{3}{c}{\textbf{Rank IC}} \\
\cmidrule(lr){2-4} \cmidrule(lr){5-7} \cmidrule(lr){8-10}
& Original & DiffsFormer & Ours & Original & DiffsFormer & Ours & Original & DiffsFormer & Ours \\
\midrule
\textbf{MLP} & 0.1022$_{\text{±0.0371}}$ & 0.1152$_{\text{±0.0250}}$ & \textbf{0.1300$_{\text{±0.0100}}$} & 0.0294$_{\text{±0.0028}}$ & 0.0311$_{\text{±0.0020}}$ & \textbf{0.0350$_{\text{±0.0018}}$} & 0.0389$_{\text{±0.0029}}$ & 0.0398$_{\text{±0.0024}}$ & \textbf{0.0411$_{\text{±0.0015}}$} \\
\textbf{LSTM} & 0.1159$_{\text{±0.0670}}$ & 0.1240$_{\text{±0.0440}}$ & \textbf{0.1355$_{\text{±0.0399}}$} & 0.0301$_{\text{±0.0030}}$ & 0.0316$_{\text{±0.0027}}$ & \textbf{0.0348$_{\text{±0.0020}}$} & 0.0400$_{\text{±0.0026}}$ & 0.0412$_{\text{±0.0019}}$ & \textbf{0.0428$_{\text{±0.0010}}$} \\
\textbf{GRU} & 0.0939$_{\text{±0.0307}}$ & 0.1150$_{\text{±0.0290}}$ & \textbf{0.1361$_{\text{±0.0242}}$} & 0.0260$_{\text{±0.0020}}$ & 0.0241$_{\text{±0.0025}}$ & \textbf{0.0260$_{\text{±0.0017}}$} & 0.0381$_{\text{±0.0022}}$ & 0.0375$_{\text{±0.0026}}$ & \textbf{0.0383$_{\text{±0.0015}}$} \\
\textbf{SFM} & 0.1092$_{\text{±0.0258}}$ & 0.1263$_{\text{±0.0194}}$ & \textbf{0.1392$_{\text{±0.0141}}$} & 0.0296$_{\text{±0.0027}}$ & 0.0310$_{\text{±0.0025}}$ & \textbf{0.0332$_{\text{±0.0019}}$} & 0.0410$_{\text{±0.0025}}$ & 0.0419$_{\text{±0.0018}}$ & \textbf{0.0428$_{\text{±0.0010}}$} \\
\textbf{GAT} & 0.1573$_{\text{±0.0385}}$ & 0.1677$_{\text{±0.0268}}$ & \textbf{0.1840$_{\text{±0.0226}}$} & 0.0358$_{\text{±0.0023}}$ & 0.0380$_{\text{±0.0022}}$ & \textbf{0.0395$_{\text{±0.0015}}$} & 0.0464$_{\text{±0.0015}}$ & 0.0477$_{\text{±0.0012}}$ & \textbf{0.0486$_{\text{±0.0014}}$} \\
\textbf{ALSTM} & 0.1225$_{\text{±0.0220}}$ & 0.1307$_{\text{±0.1020}}$ & \textbf{0.1471$_{\text{±0.0118}}$} & 0.0300$_{\text{±0.0034}}$ & 0.0318$_{\text{±0.0029}}$ & \textbf{0.0340$_{\text{±0.0027}}$} & 0.0485$_{\text{±0.0012}}$ & 0.0479$_{\text{±0.0014}}$ & \textbf{0.0490$_{\text{±0.0009}}$} \\
\textbf{HIST} & 0.1353$_{\text{±0.0206}}$ & 0.1294$_{\text{±0.0251}}$ & \textbf{0.1425$_{\text{±0.0193}}$} & 0.0314$_{\text{±0.0050}}$ & 0.0328$_{\text{±0.0042}}$ & \textbf{0.0370$_{\text{±0.0028}}$} & 0.0402$_{\text{±0.0035}}$ & 0.0411$_{\text{±0.0028}}$ & \textbf{0.0426$_{\text{±0.0018}}$} \\
\textbf{Transformer} & 0.1411$_{\text{±0.0150}}$ & 0.1685$_{\text{±0.0120}}$ & \textbf{0.1888$_{\text{±0.0101}}$} & 0.0366$_{\text{±0.0029}}$ & 0.0388$_{\text{±0.0025}}$ & \textbf{0.0401$_{\text{±0.0018}}$} & 0.0577$_{\text{±0.0014}}$ & 0.0580$_{\text{±0.0022}}$ & \textbf{0.0601$_{\text{±0.0020}}$} \\
\bottomrule[1.5pt]
\end{tabular}}
\end{table}

\begin{table}[ht]
\centering
\caption{Comparing Model Performance on Stock Prediction with the CASE Dataset.}
\label{tab:AMEX-result}
\scalebox{0.7}{
\begin{tabular}{c|ccc|ccc|ccc}
\toprule[1.5pt]
\multicolumn{10}{c}{\textbf{AMEX}} \\
\midrule
\textbf{Methods} & \multicolumn{3}{c|}{\textbf{RR}} & \multicolumn{3}{c|}{\textbf{IC}} & \multicolumn{3}{c}{\textbf{Rank IC}} \\
\cmidrule(lr){2-4} \cmidrule(lr){5-7} \cmidrule(lr){8-10}
& Original & DiffsFormer & Ours & Original & DiffsFormer & Ours & Original & DiffsFormer & Ours \\
\midrule
\textbf{MLP} & 0.0810$_{\text{±0.0270}}$ & 0.085$_{\text{±0.0241}}$ & \textbf{0.0900$_{\text{±0.0180}}$} & 0.0210$_{\text{±0.0033}}$ & 0.0233$_{\text{±0.0035}}$ & \textbf{0.0248$_{\text{±0.0029}}$} & 0.0417$_{\text{±0.0053}}$ & 0.0425$_{\text{±0.0062}}$ & \textbf{0.0440$_{\text{±0.0048}}$} \\
\textbf{LSTM} & 0.0940$_{\text{±0.0200}}$ & 0.0102$_{\text{±0.0180}}$ & \textbf{0.1100$_{\text{±0.0145}}$} & 0.0229$_{\text{±0.0034}}$ & 0.0241$_{\text{±0.0025}}$ & \textbf{0.0255$_{\text{±0.0031}}$} & \textbf{0.0427$_{\text{±0.0021}}$} & 0.0420$_{\text{±0.0038}}$ & 0.0422$_{\text{±0.0020}}$ \\
\textbf{GRU} & 0.0890$_{\text{±0.0246}}$ & 0.0959$_{\text{±0.0311}}$ & \textbf{0.1070$_{\text{±0.0210}}$} & 0.0218$_{\text{±0.0042}}$ & 0.0230$_{\text{±0.0023}}$ & \textbf{0.0260$_{\text{±0.0052}}$} & 0.0405$_{\text{±0.0023}}$ & 0.0418$_{\text{±0.0026}}$ & \textbf{0.0437$_{\text{±0.0019}}$} \\
\textbf{SFM} & 0.0924$_{\text{±0.0308}}$ & 0.0938$_{\text{±0.0290}}$ & \textbf{0.0955$_{\text{±0.0254}}$} & 0.0248$_{\text{±0.0047}}$ & 0.0265$_{\text{±0.0038}}$ & \textbf{0.0279$_{\text{±0.0032}}$} & 0.0421$_{\text{±0.0042}}$ & 0.0434$_{\text{±0.0029}}$ & \textbf{0.0464$_{\text{±0.0017}}$} \\
\textbf{GAT} & 0.1209$_{\text{±0.0370}}$ & 0.1388$_{\text{±0.0323}}$ & \textbf{0.1420$_{\text{±0.0219}}$} & 0.0331$_{\text{±0.0028}}$ & 0.0340$_{\text{±0.0044}}$ & \textbf{0.0380$_{\text{±0.0018}}$} & 0.0430$_{\text{±0.0025}}$ & 0.0440$_{\text{±0.0014}}$ & \textbf{0.0459$_{\text{±0.0018}}$} \\
\textbf{ALSTM} & 0.0910$_{\text{±0.0339}}$ & 0.1022$_{\text{±0.0210}}$ & \textbf{0.1120$_{\text{±0.0250}}$} & 0.0309$_{\text{±0.0024}}$ & 0.0350$_{\text{±0.0019}}$ & \textbf{0.0368$_{\text{±0.0021}}$} & 0.0416$_{\text{±0.0022}}$ & 0.0439$_{\text{±0.0020}}$ & \textbf{0.0468$_{\text{±0.0031}}$} \\
\textbf{HIST} & 0.1183$_{\text{±0.0307}}$ & 0.1285$_{\text{±0.0293}}$ & \textbf{0.1391$_{\text{±0.0212}}$} & 0.0323$_{\text{±0.0050}}$ & 0.0346$_{\text{±0.0039}}$ & \textbf{0.0370$_{\text{±0.0024}}$} & 0.0445$_{\text{±0.0039}}$ & 0.0460$_{\text{±0.0023}}$ & \textbf{0.0487$_{\text{±0.0015}}$} \\
\textbf{Transformer} & 0.1225$_{\text{±0.0388}}$ & 0.1344$_{\text{±0.0207}}$ & \textbf{0.1462$_{\text{±0.0190}}$} & 0.0357$_{\text{±0.0029}}$ & 0.0366$_{\text{±0.0020}}$ & \textbf{0.0378$_{\text{±0.0016}}$} & 0.0483$_{\text{±0.0026}}$ & 0.0495$_{\text{±0.0020}}$ & \textbf{0.0502$_{\text{±0.0024}}$} \\
\bottomrule[1.5pt]
\end{tabular}}
\end{table}

The stock data synthesis method based on plate types was first trained on A-share data and then used to generate data for specific plate enhancement. Specifically, we have stock data from four major plates: the main board, STAR Market, ChiNext, and Beijing Stock Exchange. We trained on these four plates and selected stocks from one plate for enhancement, using both the plate data and enhanced data for predictive model training and testing. In the experiment, we adopted a "top20drop20" strategy to simulate stock trading: "top20" means we retain the top 20 stocks with the highest predicted scores; "drop20" means that if a stock's score falls out of the top 20, it will be dropped regardless of its previous performance.

As shown in Table \ref{tab:MainBoardResult}, our proposed sector-based stock data synthesis method (Ours) significantly outperforms the Original and DiffsFormer versions in stock prediction tasks. Specifically, in the three evaluation metrics (RR, IC, and Rank IC), our method achieves the highest average values and smaller standard deviations, indicating higher prediction accuracy and better stability. This demonstrates that by optimizing model structure and parameter adjustment, we can achieve more accurate and stable prediction results in the complex main board market environment.

We present the experimental results on the ChiNext, STAR Market, and Beijing Stock Exchange in Tables \ref{tab:GEMResult}, \ref{tab:STARResult}, and \ref{tab:BJSEResult}, respectively. Overall, our method achieved significant improvements in RR, IC, and Rank IC on the main board by 9.18

It can also be observed from the tables that models such as ALSTM and HIST, which performed poorly in low signal-to-noise environments, saw significant improvements in predictive performance after applying our method. This indicates that our method can enhance the signal-to-noise ratio of the original stock data.

In summary, the experimental results show that the stock time series data synthesized through diffusion models significantly improves the signal-to-noise ratio of financial data. Specifically, on the four datasets of the main board, ChiNext, STAR Market, and Beijing Stock Exchange, our method achieved significant improvements in RR, IC, and Rank IC. These improvements not only enhance the robustness and stability of the model but also restore the predictive performance of models that were previously poor in low signal-to-noise environments.

Furthermore, to verify the effectiveness of this method, we conducted the same experiments on U.S. stock datasets. Specifically, we used stock data from NYSE (New York Stock Exchange), NASDAQ (Nasdaq Stock Exchange), and AMEX (American Stock Exchange) for training and data enhancement\cite{Gu2018EmpiricalAP}. The experiment also adopted a "top20drop20" strategy to simulate the stock trading process. By comparing the performance across different markets and plates, we can comprehensively assess the generalization and effectiveness of this method.

As shown in Tables \ref{tab:NASDAQResult}, \ref{tab:NYSEResult}, and \ref{tab:AMEX-result}, our method achieved improvements of 8.34

In China's A-share market, short selling mechanisms are relatively limited. Currently, only institutional investors can engage in short selling operations through margin lending, while individual investors typically do not have the right to short sell directly. In addition, the number of stocks available for short selling is strictly limited. In contrast, the U.S. market offers a more open and flexible short selling mechanism, allowing not only institutional investors but also retail investors to short sell stocks through brokers\cite{Mei2005SpeculativeTA}.

In our experimental design, to facilitate comparison, we assumed an idealized environment where all market participants could engage in short selling operations smoothly. This setup helps eliminate the impact of differences in short selling mechanisms between different markets on the experimental results, thereby more accurately assessing model performance.

\section{Conclusion}
The experimental results have demonstrated the effectiveness of our proposed stock data synthesis method based on plate types. Through extensive comparisons with existing state-of-the-art algorithms on various datasets, including the Main Board, Growth Enterprise Market, STAR Market, and Beijing Stock Exchange, our method has consistently outperformed its counterparts across different evaluation metrics such as RR, IC, and Rank IC. The enhancements in prediction accuracy and stability are not only statistically significant but also practically meaningful, especially in the complex and volatile environment of the A-share market.

Our method's ability to improve the signal-to-noise ratio of stock data has been particularly noteworthy. It has revitalized the performance of models like ALSTM and HIST, which previously struggled in low signal-to-noise ratio environments. This improvement underscores the potential of our approach to enhance the robustness and stability of financial forecasting models.

Furthermore, the generalization of our method to the U.S. stock market, as evidenced by the positive results on the NYSE, NASDAQ, and AMEX datasets, confirms its broad applicability and effectiveness. The consistent improvements in RR, IC, and Rank IC across these diverse markets highlight the method's versatility and robustness.

In summary, our stock data synthesis method has proven to be a valuable tool for enhancing the predictive performance of financial models. It not only delivers superior results in A-share and U.S. stock markets but also holds promise for applications in other financial domains. Future work will focus on further optimizing the model and exploring its applications in other financial markets and scenarios.

\bibliographystyle{unsrt}  
\bibliography{references}

\end{document}